\documentclass[conference]{IEEEtran}
\IEEEoverridecommandlockouts

\usepackage{cite}
\usepackage{amsmath,amssymb,amsfonts}
\usepackage{algorithmic}
\usepackage{graphicx}
\usepackage{textcomp}
\usepackage{xcolor}
\def\BibTeX{{\rm B\kern-.05em{\sc i\kern-.025em b}\kern-.08em
    T\kern-.1667em\lower.7ex\hbox{E}\kern-.125emX}}
\usepackage{subfig}
\usepackage{hyperref}
\usepackage{authblk} 
\usepackage{etoolbox} 
\usepackage[pro]{fontawesome5}

\begin{document}
\IEEEpubid{\makebox[\columnwidth]{ 979-8-3315-2710-5/25/\$31.00 ©2025 IEEE
\hfill{ \hspace{\columnsep} \makebox[\columnwidth]{ }}}}

\title{TruncQuant: Truncation-Ready Quantization for DNNs with Flexible Weight Bit Precision}

\author{
    Jinhee Kim\textsuperscript{1*}\thanks{$^{*}$: Equal contributions},
    Seoyeon Yoon\textsuperscript{2*},
    Taeho Lee\textsuperscript{2},
    Joo Chan Lee\textsuperscript{3},
    Kang Eun Jeon\textsuperscript{1\faEnvelope[regular]}\thanks{\textsuperscript{\faEnvelope[regular]}: Corresponding authors},
    Jong Hwan Ko\textsuperscript{1\faEnvelope[regular]} \\
    
    \textsuperscript{1}Department of Electrical and Computer Engineering,
    \textsuperscript{2}School of Electronic and Electrical Engineering,\\
    \textsuperscript{3}Department of Artificial Intelligence, Sungkyunkwan University, Suwon, Korea \\

    Email: {\small\texttt{\{a2jinhee,syy000405,dlxogh063,maincold2,kejeon,jhko\}@skku.edu
    }}
}

\maketitle

\begin{abstract}
The deployment of deep neural networks on edge devices is a challenging task due to the increasing complexity of state-of-the-art models, requiring efforts to reduce model size and inference latency. Recent studies explore models operating at diverse quantization settings to find the optimal point that balances computational efficiency and accuracy. Truncation, an effective approach for achieving lower bit precision mapping, enables a single model to adapt to various hardware platforms with little to no cost. However, formulating a training scheme for deep neural networks to withstand the associated errors introduced by truncation remains a challenge, as the current quantization-aware training schemes are not designed for the truncation process. We propose TruncQuant, a novel truncation-ready training scheme allowing flexible bit precision through bit-shifting in runtime. We achieve this by aligning TruncQuant with the output of the truncation process, demonstrating strong robustness across bit-width settings, and offering an easily implementable training scheme within existing quantization-aware frameworks. Our code is released at \href{https://github.com/a2jinhee/TruncQuant}{https://github.com/a2jinhee/TruncQuant}.
\end{abstract}

\section{Introduction}
\label{label:introduction}

While deep neural networks excel across various benchmarks, their computational demands typically grow with capacity, making runtime costs a crucial consideration in practice~\cite{burrello2021dory}. As a result, finding a good trade-off by balancing efficiency and accuracy has become increasingly important when deploying models on power-constrained edge devices. One of the most widely used techniques to tackle this problem is quantization, which represents full-precision weights as reduced-precision integers and in turn significantly lowers the model’s memory and computational demands~\cite{yang2019quantization, banner2019post, nagel2020up, li2021brecq}. 

Beyond quantization, various approaches have been proposed to optimize the trade-off for greater efficiency. Active research areas include adaptive inference strategies~\cite{yu2018slimmable, yu2019universally, tan2019efficientnet} and dynamically controlled termination methods~\cite{teera2016branchynet,laskaridis2021adaptive}, both of which help reduce runtime power consumption by finding the right balance between accuracy and efficiency. These efforts are also recognized alongside a plethora of hardware-related studies~\cite{ryu2022review,camus2019review} that realize low-power accelerators with scalable, fine-grained control~\cite{sharma2018bit,moons2017envision, ryu2019bitblade,tahmasebi2024flexibit}. However, previous methods require identifying an optimal operation point, resulting in the need to create dedicated models for each specific constraint- whether for deployment across different platforms or to handle changing resource availability on a single device.

To address this issue, several works propose to train a single model capable of supporting multiple quantization precisions~\cite{xu2023eq,chmiel2020robust,xu2022multiquant,zhong2023mbquant}. In such flexible networks, or Once-For-All (OFA) networks, multiple child models of reduced precisions are generated from a single full-precision parent model. Because the quantized child models share the parent model, the storage overhead for hosting multiple dedicated models is neutralized to some extent. 
Nevertheless, the parent model must still reside in host memory for reference. Additionally, generating quantized models introduces extra floating-point operations from the quantization function and requires costly memory accesses—both to fetch the floating-point model and to store the quantized model in slow off-chip memory~\cite{dao2022flashattention}.

To overcome such shortcomings, we propose the use of truncation, through which a low precision model can be attained by merely discarding the least significant bits (LSB); in other words, the most significant bits (MSB) of the quantized weights are shared. Perhaps the most related work would be Any-Precision \cite{yu2021any}, as it also proposes bit-shifting to realize flexible bit precision of weights. During training, truncation functions are typically approximated as quantization functions, with the expectation that accuracy will be preserved with bit-shifting inference. However, we discover that quantization functions fail to accurately reflect truncation operations, and lead to catastrophic accuracy drop for lower precisions. Therefore, to maintain accuracy in lower precisions, the issue remains that we have to use quantization when generating child models, requiring the large parent model to be stored in off-chip memory and retrieved each time for this process.

\begin{figure}[t]
    \centering
    \includegraphics[width=\linewidth]{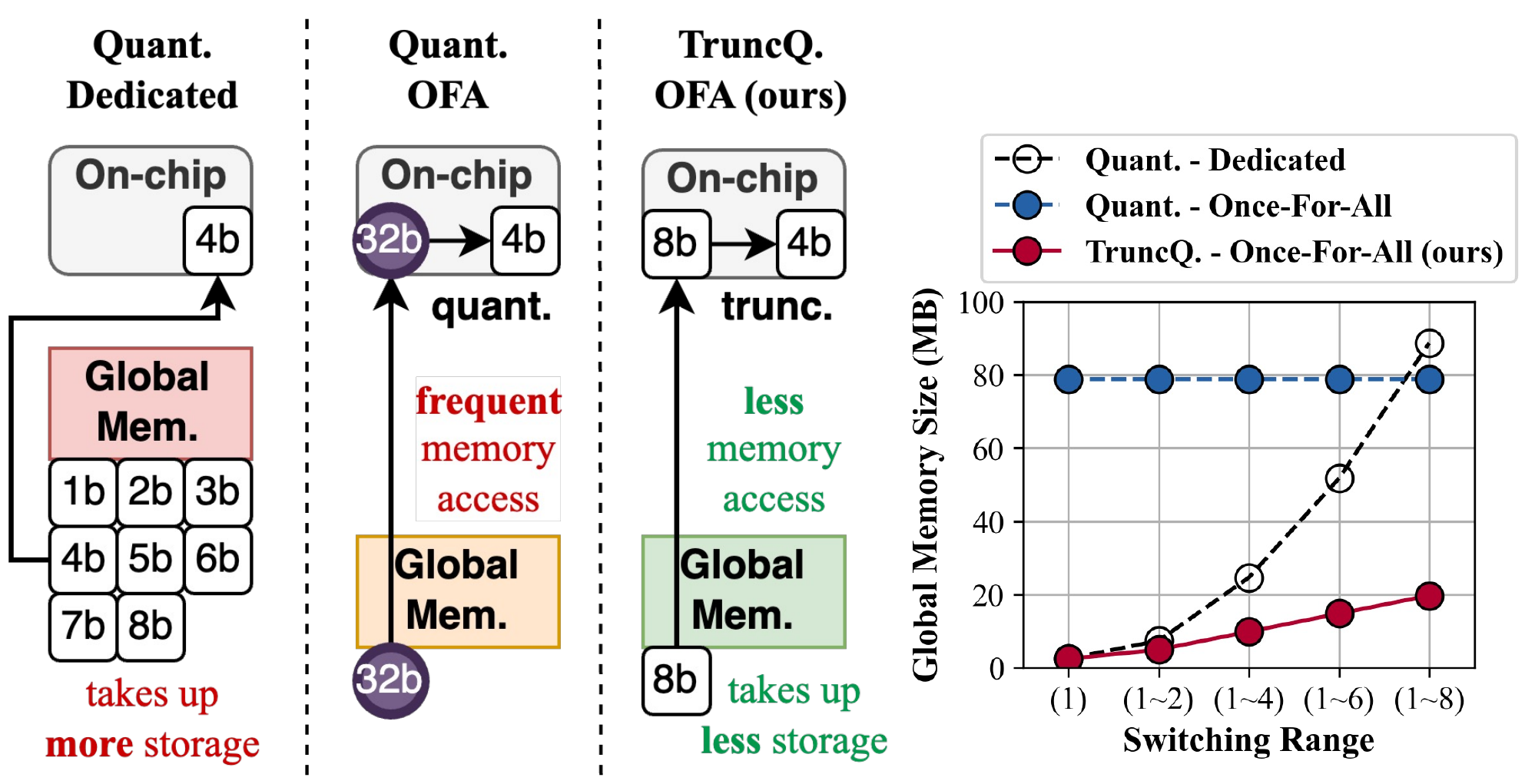}
    \caption{A conceptual diagram depicting three approaches for implementing flexible quantization: multiple dedicated models, OFA networks, and our proposed method (left), along with a comparison of their memory requirements (right).
    }
    \label{fig:storage}
    \vspace{-0.3cm}
\end{figure}

To bridge this gap between quantization and truncation operations, we propose TruncQuant, a novel truncation-ready quantization scheme that enables weights to achieve flexible bit precision solely through bit-shifting. We identify certain ranges of floating-point values that end up with different integer representation depending on the used operation-quantization or truncation. The formulation of TruncQuant builds upon this observation by addressing and eliminating this range altogether. By doing so, TruncQuant can recover the 67.74\% accuracy drop caused by simply truncating QAT-trained weights for ResNet-50 quantized to 2 bits, on ImageNet. At runtime, TruncQuant achieves approximately 3.53$\times$ storage savings for ResNet-50 \cite{yu2021any} by requiring only an 8-bit model instead of a floating-point model, as in conventional methods, while also reducing costly memory accesses when adjusting model precision.
Fig.~\ref{fig:storage} presents our method alongside existing approaches for flexible quantization, and compares their storage requirements. The comparison highlights the efficiency of our method in reducing memory usage.

To summarize, our contributions are:
\begin{itemize}
    \item We first identify the existence of quantization-truncation error and present detailed formulations to characterize it.
    \item Based on our above findings, we introduce TruncQuant, a truncation-ready quantization scheme,  which accurately reflects the impact of truncation on the deep learning model performance. 
    \item Lastly, we validate the effectiveness of TruncQuant by demonstrating accuracy on par with the baseline and hardware benefits in terms of storage and energy savings.
\end{itemize}

\section{Preliminaries}
\label{label:preliminaries}

\noindent\textbf{Quantization.} Quantization maps continuous real-valued weights to a discrete set of numbers with reduced numerical precision, thereby accelerating computation and saving storage~\cite{gholami2022survey, qin2020binary}. In this work, we focus on uniform quantization \cite{jacob2018quantization}, the most widely adopted quantization scheme due to its simplicity and effectiveness in hardware implementation. 
An $n$-bit uniform quantizer, which maps a floating-point weight $\mathbf{W} \in \mathbb{R}$ to its quantized counterpart $\mathbf{\bar{Q}}_n \in \mathbb{Z}_n=\{0,1,\dots, 2^n-1\}$, is expressed as follows:
\begin{align} \label{eq:quant_op} 
&q(\mathbf{W}; n) = \left\lceil \dfrac{\mathbf{W}}{\Delta_n} \right\rfloor + z_n,
\end{align}
where $\Delta_n$ and $z_n$ are the quantization step size and the zero point for target precision $n$; and $\left\lceil \cdot \right\rfloor$ is the rounding function. 

Quantization error is then defined as the distance between the original weights and the quantized ones \cite{esser2019learned} summed across all quantized layers as shown below:
\begin{align}\label{eq:quant_error} 
&E_Q = \sum_{\forall l} \|{\mathbf{W}}^{(l)}-\mathbf{Q}_n^{(l)}\|,
\end{align}
where $\mathbf{Q}_n^{(l)}=\Delta_n(\mathbf{\bar{Q}}^{(l)}_n-z_n)$ represents the dequantized weight of layer $l$; and $\|\cdot\|$ is an arbitrary type of distance function (\textit{e.g.}, $L^p$ norm). 

\begin{figure}[t]
  \centering
  \includegraphics[width=0.95\columnwidth]{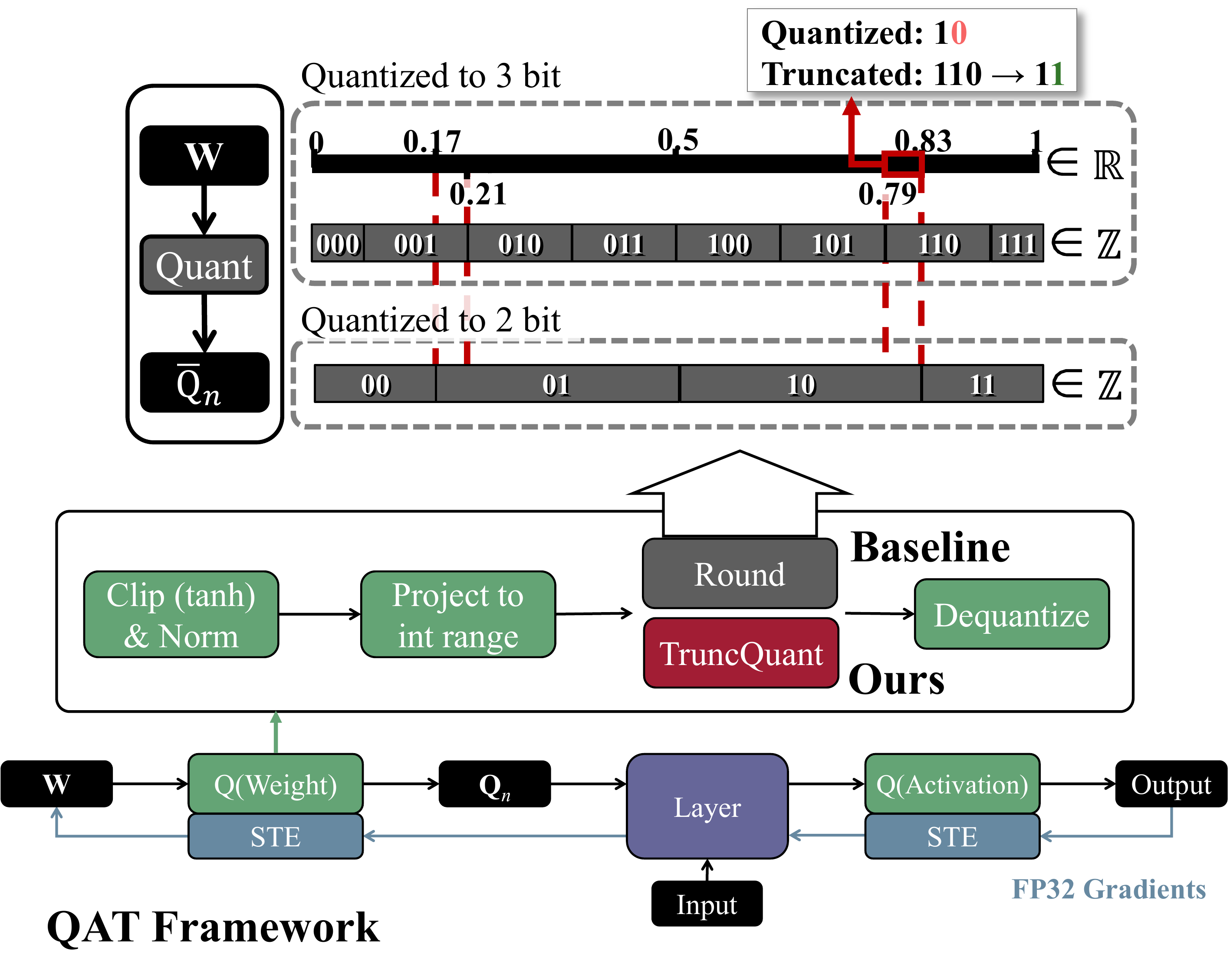}
  \caption{
  Summary of the QAT framework and how our method integrates within the weight quantizer.
  In baseline methods, uniform quantization and truncation yields different results for certain floating-point ranges. 
  }
  \label{fig:qt_op}
\end{figure}

Naturally, lower quantization precision inevitably yields higher quantization errors, often resulting in significant model performance degradation. 
A common approach to address this issue is to simulate the quantization scheme during training \cite{jacob2018quantization}. 
As shown in Fig.~\ref{fig:qt_op}, the quantization-aware-training (QAT) framework \cite{yu2021any, chmiel2020robust, xu2023eq, shen2021once} quantizes weights during the forward pass, but maintains floating-point value gradients using the straight-through estimator~\cite{ste}. 

\vspace{1em}
\noindent\textbf{Truncation.} Truncation offers an efficient and cost-free method to achieve the desired model precision by simply removing the least significant bits (LSBs) of the weights.
When $b$ is the precision the model begins with (starting bit precision for truncation), and $n$ is the target bit precision, truncation from $b$-bit to $n$-bit integer representation $t(\cdot; b, n):\mathbb{Z}_b\rightarrow \mathbb{Z}_n$ can be expressed as follows:
\begin{equation} \label{eq:trunc_op} 
t(\mathbf{\bar{Q}}_b;b, n) = \mathbf{\bar{Q}}_b \gg (b-n),
\end{equation}
where $b>n$; and $\mathbf{\bar{T}}_n=t(\mathbf{\bar{Q}}_b;b, n) $ is defined as the truncated integer representation from a higher-bit tensor. 

Uniform quantization and truncation, in essence, is mapping each weight to a specific ‘\textit{bin}’ within a defined range of integer values. As shown in Fig.~\ref{fig:qt_op}, this bin can be acquired by either directly rounding the scaled value on the integer range (quantization), or bit-shifting from a higher bit to a specified bit precision integer range (truncation). Both methods rely on \textit{quantization/truncation boundaries} that divide the range of floating-point values into different bins. 

\begin{figure*}[t]
    \centering
    \includegraphics[width=0.8\linewidth]{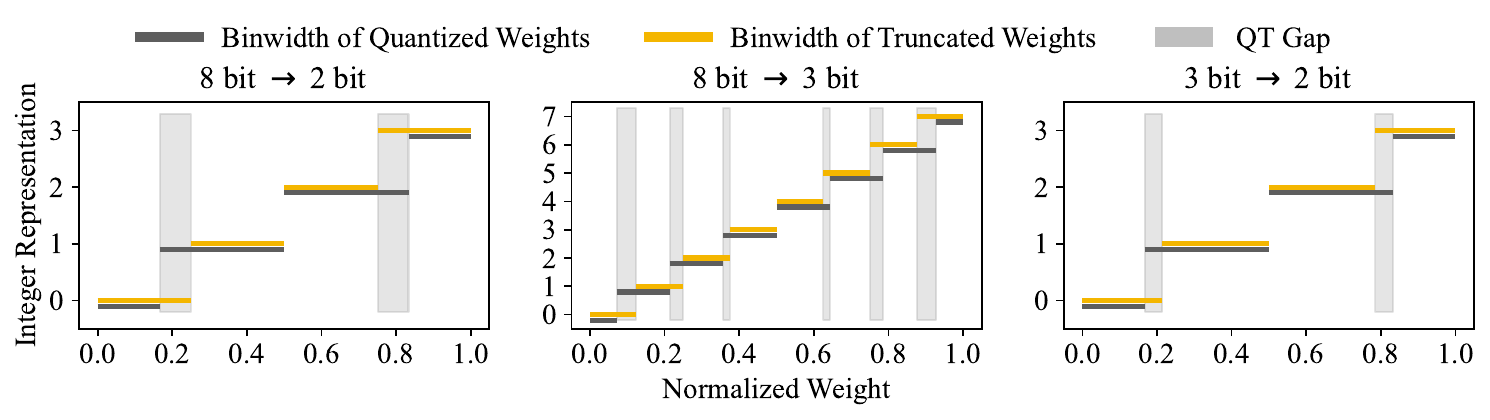}
    \caption{Three quantization-truncation scenarios illustrating the QT Gap (shaded in grey), where the weight is quantized to $b$-bit and then truncated to $n$-bit: ${b}=8$ and $n=2$ (left), ${b}=8$ and $n=3$ (middle), ${b}=3$ and $n=2$ (right). }
    \label{fig:qt_gap}
    \vspace{-0.3cm}
\end{figure*}

\section{Quantization-Truncation Error}
\label{sec:truncation_error}

We first identify the innate differences between the quantization and truncation operation and the accompanying error. To measure this error, we define \textbf{Quantization-Truncation Error (QT Error)} as the distance between the uniform quantized weights and truncated weights with respect to a specific bit precision. The QT Error is expressed as follows:
\begin{align}\label{eq:trunc_error}
E_T = \sum _{\forall l} \|\mathbf{T}_n^{(l)}-\mathbf{Q}_n^{(l)}\|,
\end{align}
where $\mathbf{T}_n^{(l)}, \mathbf{Q}_n^{(l)}$ is obtained by scaling $\mathbf{\bar{T}}_n, \mathbf{\bar{Q}}_n$ back to the original floating-point range for layer $l$.

As shown in Fig.~\ref{fig:qt_gap}, certain floating-point ranges can lead to different bin assignments under quantization and truncation, contributing to the QT Error that compounds with repeated truncation. Two key factors influence the QT Error: \textbf{i) the Quantization-Truncation Gap (QT Gap)} defines the range of values that have different bin assignments, and \textbf{ii) the level size} shows the amount of impact this misalignment has on the QT Error. Together, these two factors shape the observed QT Error trend in Fig.~\ref{fig:weightsQT} (top), where it peaks at 2-bits and decreases with higher precision.

Therefore, the QT Error can be restated as the product of the number of weights within the QT Gap and the level size: 
\begin{align}
\label{eq:trunc_error_with_QT}
E_T = \Delta^{'} s_n \sum _{\forall l} N^{(l)},
\end{align}
where $N^{(l)}$ is the number of weights in the QT Gap for layer $l$; and the level size $s_n$ and the miscellaneous scaling factor $\Delta^{'}$ make up the quantization step size $\Delta = \Delta^{'}s_n$. For the DoReFa quantizer, $\Delta^{'}=2\mathbb{E}(|\mathbf{W}|)$ \cite{zhou2016dorefa}. 

\subsection{Quantization-Truncation Gap (QT Gap)}
\label{sec:qt_gap}

Our initial objective is to pinpoint the boundaries where floating-point values map to different bins under quantization and truncation. When target precision $n$ is specified, the maximum bin value is $M_n=2^n-1$, assuming that the inputs are normalized to [0,1] before the process. To obtain the quantized bins, we first scale the original value to the range of $[0, M_n]$, expressed in floating-point format. This is then followed by rounding to the nearest integer value. However, to obtain the truncated bins, we bit-shift from a higher precision integer range to the desired integer range of $[0, M_n]$. As a practical example, consider truncation starting from 8-bit integers (INT8). To obtain the truncated bin, we first scale and round the floating-point value to the high precision integer range $[0, M_8]$, and bit-shift to the target range $[0, M_n]$.

\begin{figure}[t]
    \captionsetup[subfigure]{justification=centering}
    \centering
    \includegraphics[width=0.75\linewidth]{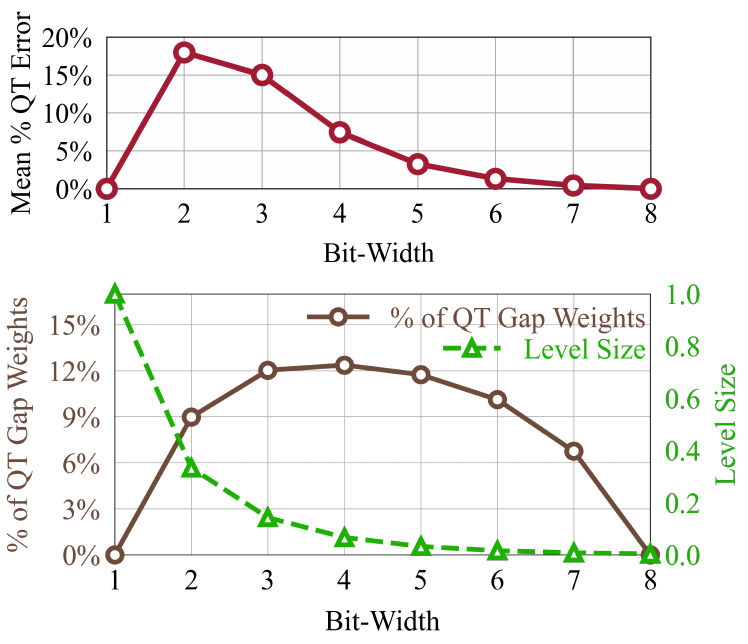}    
    \caption{The QT Error trend can be explained as the combined trend of the number of weights in the QT Gap and the level size. We use $L^1$ norm to plot the QT Error.}
    \label{fig:weightsQT}
    \vspace{-0.3cm}
\end{figure}

Despite their shared goal of mapping values to a specific integer, quantization and truncation can assign different bins to certain weights. To identify weights susceptible to this discrepancy, we compute the `binwidth’ which groups the range of floating-point values that map to the same bin under each method. The binwidth of quantized and truncated weights for the uniform quantizer can be expressed as follows:
\begin{align} 
\label{eq:quant_domain} 
& R_Q(i, n) = \Big[\frac{i-0.5}{M_n}, \frac{i+0.5}{M_n}\Big) \\[1pt]
\label{eq:trunc_domain}
& R_T(j, n) = \Big[\frac{j \cdot 2^{{b}-n} - 0.5}{M_{{b}}}, \frac{(j+1) \cdot 2^{{b}-n} - 0.5}{M_{{b}}}\Big),
\end{align}
where $\forall i,j=\{0,1,...,M_n\}$; and ${b}$ is the arbitrary starting bit precision for truncation. In order for the quantized and truncated bin to be $i, j$, the elements of the normalized floating-point weight tensor $\mathbf{W}^{'}$ need to be in the range $R_Q(i,n)$ and range $R_T(j,n)$, respectively. Interestingly, as shown in Fig.~\ref{fig:weightsQT} (bottom), the number of weights in the QT Gap peaks in the middle bit-width range (e.g., 4 bits). This reflects the interplay between two opposing trends: higher precision introduces more potential gaps, while simultaneously reducing the size of each gap.

\subsection{Level Size}
Weights in the QT Gap are assigned to different bins, which typically differ by only one step. The resulting difference is determined by the scaling factor used to convert these bins back to the floating-point range. To quantify this difference, we introduce the level size, denoted as $s_n = 1/M_n=1/(2^n-1)$ where $M_n$ is the maximum integer value of quantization levels existing in the specified bit precision $n$. This factor captures the magnitude of the mapping difference for each weight within the QT Gap, providing a measure of how much each weight contributes to the overall QT Error trend. The level size is inversely proportional to $M_n$, as shown in Fig.~\ref{fig:weightsQT} (bottom). To minimize QT Error, our proposed method aims to make the QT Gap non-existent as opposed to controlling the level size, which is an inherent property of the quantizer.

\section{TruncQuant}
We propose TruncQuant, a truncation-ready quantization scheme that enables weights to achieve flexible bit precision solely through bit-shifting inference. The main objective of the function is to address the QT Error by aligning the binwidth of quantized and truncated weights to be the same. 
Moreover, this function should be independent of the starting bit precision, as truncation may not always begin at 8 bits, especially when limited on-chip buffer size prevents storing such a model. 
To meet these two key objectives, we design the quantized binwidths to follow a multiplicative relationship—for instance, obtaining a 3-bit binwidth by halving each 2-bit binwidth—ensuring that the binwidths are evenly distributed. This guarantees that the MSBs of lower-precision weights remain aligned with those of higher-precision counterparts.

This approach can be understood through an alternative formulation of truncation. In addition to Eq.~\ref{eq:trunc_op}, when truncating from an initial $b$-bit precision to a target $n$-bit precision, the operation can be expressed as:
\begin{equation} \label{eq:trunc_op1} 
t(\mathbf{\bar{Q}}_b;b, n) = \left\lfloor \dfrac{\mathbf{\bar{Q}}_b}{2^{(b-n)}} \right\rfloor,
\end{equation}
where $b>n$; and $\left\lfloor \cdot \right\rfloor$ is the element-wise floor function. Removing the LSBs corresponds to a bitwise shift, meaning that the target precision is obtained by dividing by powers of two. Hence, by ensuring that quantized binwidths follow a structured multiple relationship, we guarantee that the MSBs remain consistent across different precision levels.

\begin{figure}[t]
    \centering
    \includegraphics[width=0.85\linewidth]{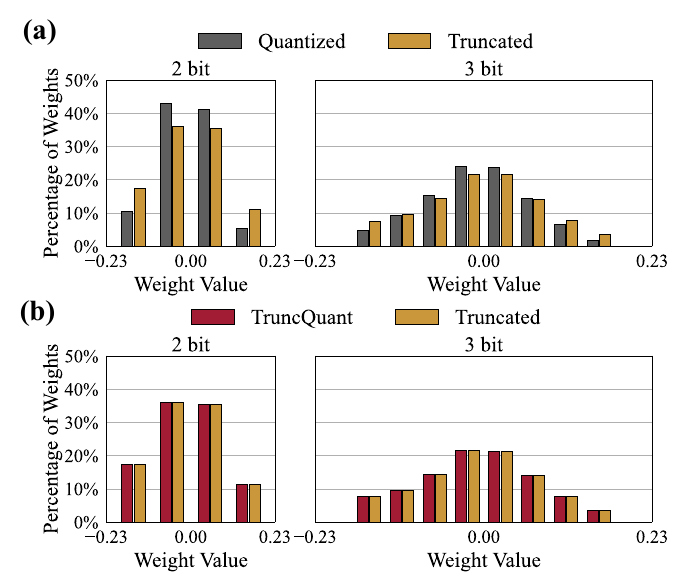}
    \caption{Weight distribution of the first quantized CONV layer in ResNet-20. Comparison between: (a) uniform quantization - truncation, and (b) TruncQuant - truncation.}
    \label{fig:conv_dist}
    \vspace{-0.3cm}
\end{figure}
\begin{table*}[t]
    \centering
    \caption{Accuracy for baseline and TruncQuant models for CIFAR-10, and SVHN.}
    \includegraphics[width=0.85\linewidth]{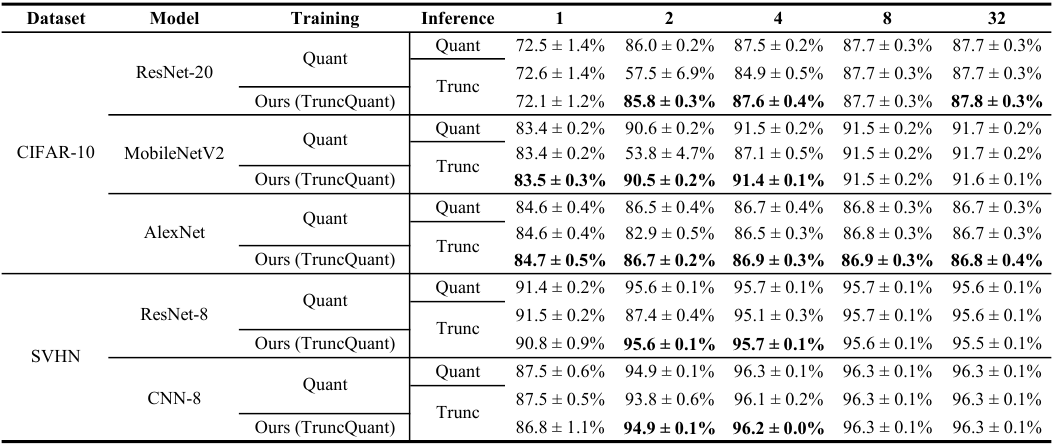}
    \label{table:eval_truncquant_table}
    \vspace{-0.3cm}
\end{table*}
The binwidth of truncated weights for uniform quantizers is divided into $(M_n+1)$ ranges, where $M_n=2^n-1$. These ranges are not equal as can be seen in Eq.~\ref{eq:trunc_domain}. To make the truncated weight binwidth of TruncQuant to be evenly distributed into $(M_n+1)$ ranges, we leverage the floor function, represented as $\left\lfloor \cdot \right\rfloor$, after scaling the input by $(M_n+1)$. This enforces our binwidth of truncated weights to be evenly distributed. Given a floating-point weight $\mathbf{W} \in \mathbb{R}^{w\times h}$, the $n$-bit TruncQuant function can be represented as follows:
\begin{align} \label{eq:truncquant}
&q_t(\mathbf{W}; n) = \left\lfloor (M_n+1)\mathbf{W} \right\rfloor,
\end{align}
where $\mathbf{W}$ is the original floating-point tensor; and $\left\lfloor \cdot \right\rfloor$ is the element-wise floor function. In Fig.\ref{fig:conv_dist}(a), we quantize the weights following the traditional QAT scheme, truncate them using bit-shifting inference, and observe a clear mismatch between the two distributions at low bit settings (e.g., 2, 3 bits). However, in Fig.\ref{fig:conv_dist}(b), we see no difference between the weights quantized with TruncQuant and the truncated weights.

It is worth noting that the binwidth of quantized and truncated weights for TruncQuant will both follow the simple range shown below: 
\begin{align} \label{eq:ideal_trunc_domain}
& R_E(k, n) = \Big[\frac{k}{M_n + 1}, \frac{k+1}{M_n + 1}\Big), 
\end{align}
where $\forall k =\{0,1,...,M_n\}$. We set this to be the \textit{truncation-ready} binwidth, as it follows the truncation behavior while being independent of the starting bit precision. This truncation-ready binwidth serves as the conceptual basis of TruncQuant. 

Following this observation, we can modify the straight-through-estimator (STE) to reflect TruncQuant's truncation-ready binwidth. Originally, the STE approximates the quantizer with a linear function that maps the full input range $[0,1]$ to the binwidth centers—that is, it uses an affine function to allow passing gradients despite the true gradients being zero. In many quantization studies, the STE is taken as the identity, or piecewise adaptive functions~\cite{ste,liu2022nonuniform,yin2019understanding}. For TruncQuant, since the binwidths are defined with a width of $1/(M_n+1)$ instead of the original $1/M_n$, the STE should be adjusted as follows: 
\begin{equation} \label{eq:ste}
    \frac{\partial \mathcal{L}}{\partial \mathbf{W}} = \frac{M_n}{M_n + 1}\frac{\partial \mathcal{L}} {\partial \mathbf{\bar{Q}}},
\end{equation} 
where the gradient update during the backward pass is scaled by a factor of $M_n/(M_n+1)$ with respect to the incoming loss $\mathcal{L}$. 
While this adjustment improves theoretical consistency, empirical results show that the original STE also performs well with TruncQuant, suggesting that the modification may not be strictly necessary in practice.

\begin{table*}[t]
    \centering
    \caption{Accuracy for baseline and TruncQuant models for ImageNet.}
    \includegraphics[width=0.85\linewidth]{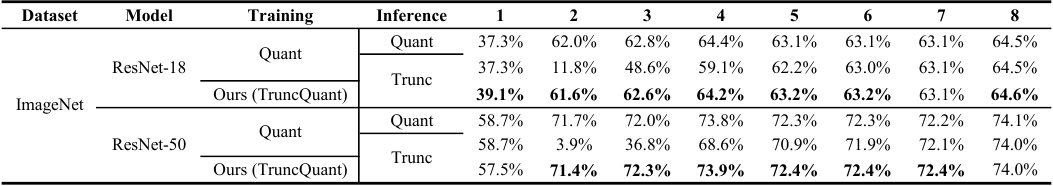}
    \label{table:truncquant_imagenet}
\end{table*}

\section{Experiments}
\label{sec:experiments}

\subsection{Setups}
\label{sec:setups}
\noindent\textbf{Baseline and experiment settings.} 
We modify the QAT framework by replacing its rounding-based quantizer with our proposed TruncQuant and compare with conventional QAT schemes (mainly \cite{yu2021any}). The na\"ive baseline is to directly truncate the quantization-aware-trained weights by simply cutting off the LSBs. For instance, to acquire an $n$-bit model, we perform bit-shifting until the target bit precision $n$. We do this for all weights, except for those in the first and last layers, following DoReFa~\cite{zhou2016dorefa}. We calibrate the model to be flexible to varying precisions following Any-Precision's method~\cite{yu2021any}.

\begin{figure}[t]
    \centering
    \includegraphics[width=0.9\columnwidth]{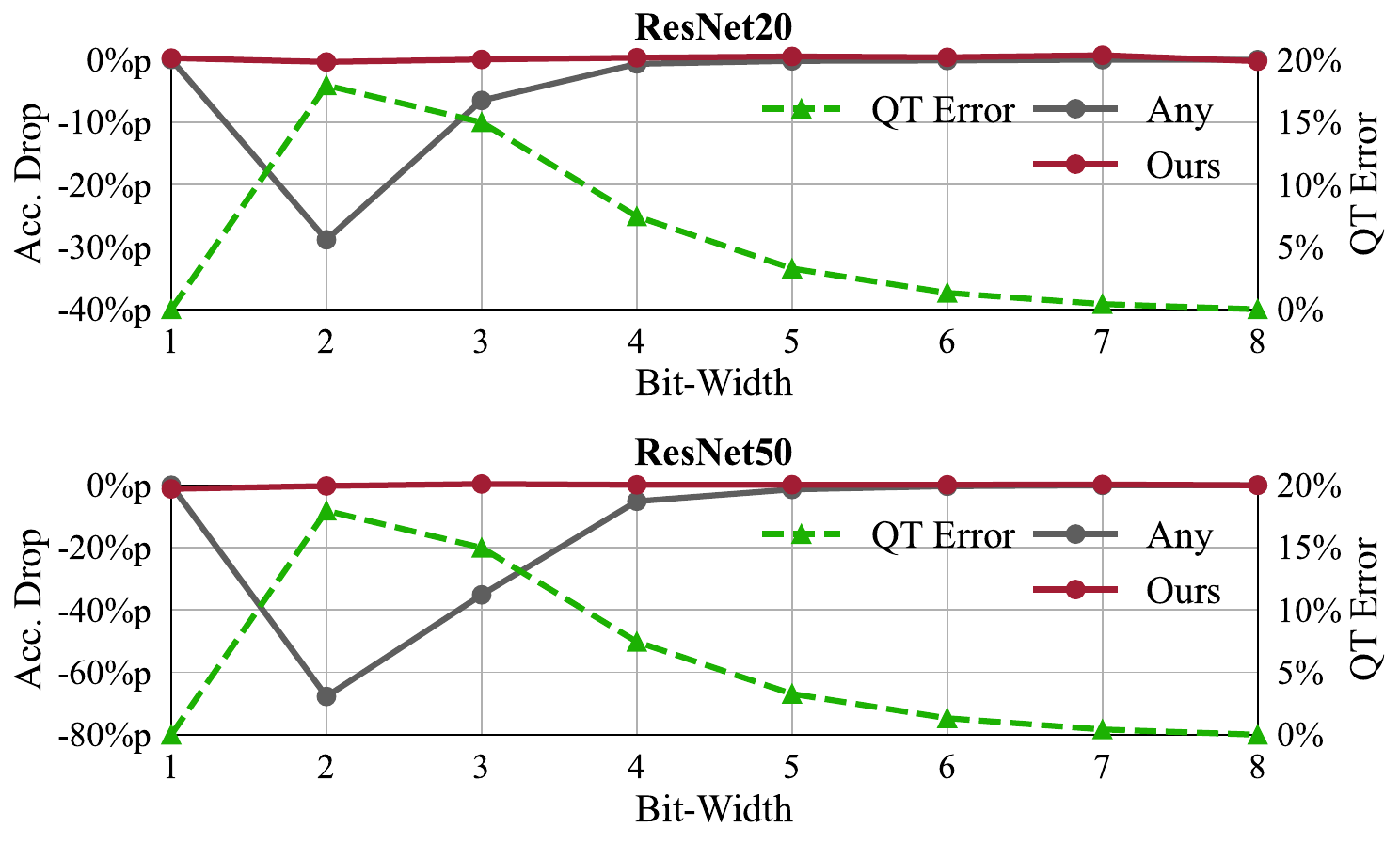}
    \caption{Comparison of accuracy drop and the QT Error of baseline and TruncQuant model (Table~\ref{table:eval_truncquant_table},~\ref{table:truncquant_imagenet}).}
    \label{fig:total_accuracy}
    \vspace{-0.3cm}
\end{figure}
\begin{table*}[t]
    \centering
    \caption{TruncQuant integration on other QAT frameworks.}
    \includegraphics[width=0.8\linewidth]{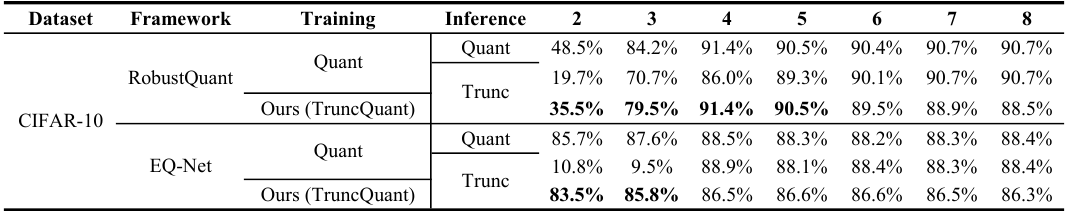}
    \label{table:truncquant_pact_eq}
    \vspace{-0.3cm}
\end{table*}

\vspace{1em}
\noindent\textbf{Models and datasets.} We validate our method with several network architectures and datasets that are well-known benchmarks for flexible quantization. These networks include ResNet-8, -18, -20, -50 \cite{he2016deep}, MobilenetV2 \cite{sandler2018mobilenetv2}, AlexNet \cite{krizhevsky2012imagenet}, and CNN-8 \cite{zhou2016dorefa}, and datasets include CIFAR-10 \cite{krizhevsky2009learning}, ImageNet \cite{deng2009imagenet}, and SVHN \cite{netzer2011reading}.

\subsection{Evaluation on CIFAR-10, SVHN, and ImageNet}
\label{sec:eval_cifar}
We demonstrate that TruncQuant preserves truncation accuracy across different bit precisions and is effective on the CIFAR-10, SVHN, and ImageNet datasets. The results are summarized in Table~\ref{table:eval_truncquant_table} and \ref{table:truncquant_imagenet}. First, we present the quantization accuracy of QAT-trained weights (Training: Quant. – Inference: Quant.). Next, we evaluate the accuracy of QAT-trained weights when directly truncated (Training: Quant. – Inference: Trunc.). Finally, we assess truncation accuracy for weights trained with TruncQuant (Training: Ours – Inference: Trunc.). Bold text in the tables highlights cases where TruncQuant outperforms the baseline in terms of truncation accuracy.

We first experiment on CIFAR-10 and SVHN with networks trained with TruncQuant using the train-from-scratch method. Table~\ref{table:eval_truncquant_table} shows that networks trained with TruncQuant demonstrate truncation accuracy on par with the original baseline. In comparison, simply truncating the baseline models leads to drastic accuracy degradation in low bit precisions (e.g. 2 bit). 
We further evaluate TruncQuant's performance on ImageNet with networks fine-tuned and calibrated using the pretrained Any-Precision model \cite{yu2021any}. 
Table~\ref{table:truncquant_imagenet} shows similar results, where weights fine-tuned with TruncQuant recover the accuracy drop observed when truncating the baseline model. In Fig.~\ref{fig:total_accuracy}, we can see that for both ResNet-20 and ResNet-50, the QT Error is inversely proportional to the accuracy drop observed when the QAT-trained model is simply truncated.

\subsection{Evaluation on other OFA frameworks}
\label{sec:eval_ofa}
In addition to our comparison with Any-Precision, we also integrate TruncQuant with other OFA frameworks such as RobustQuant \cite{chmiel2020robust} and EQ-Net \cite{xu2023eq}. The results are summarized in Table~\ref{table:truncquant_pact_eq}.
With EQ-Net, we modify the LSQ quantizer \cite{esser2019learned} to learn the scaling factor for the highest precision (e.g., 8 bit). We then derive the scaling factors for other precisions by dividing the highest precision scaling factor by powers of two. We exclude 1 bit quantization, as it is not supported in their training scheme.
Table~\ref{table:truncquant_pact_eq} shows that our proposed method recovers most of the accuracy drop in low bit precisions, and also integrates well with existing QAT frameworks.

\begin{figure}[t]
    \includegraphics[width=0.95\linewidth]{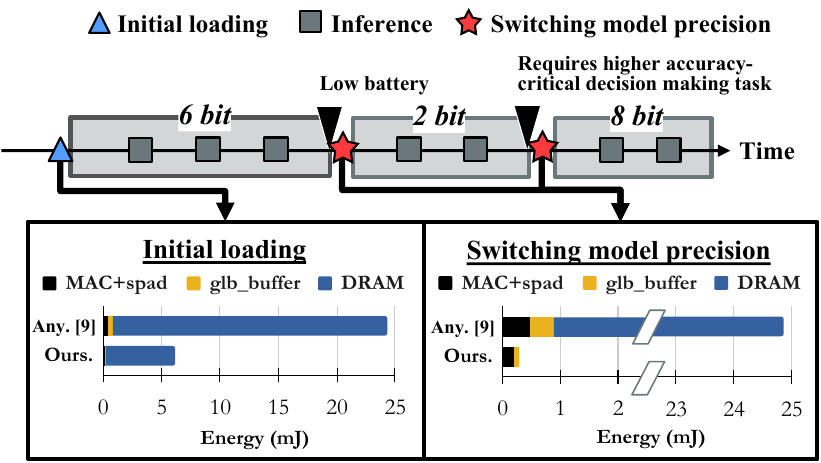}
    \caption{Runtime energy breakdown for initial loading and bit-switching scenarios for Any-Precision \cite{yu2021any} and TruncQuant.}
    \label{fig:comp}
    \vspace{-0.3cm}
\end{figure}
\subsection{Hardware implications of TruncQuant}
\label{sec:hwimp_tq}
We evaluate the energy efficiency of TruncQuant by measuring the runtime energy breakdown during inference. We compare TruncQuant against a conventional once-for-all method (i.e. Any-Precision \cite{yu2021any}), using a modified Timeloop simulator \cite{parashar2019timeloop} that supports multi-bit quantization. Specifically, we estimate energy consumption in two scenarios: 1) initial model loading and 2) switching bit precision during inference.

A key distinction is that Any-Precision requires fetching the floating-point model into the on-chip global buffer, which leads to costly DRAM accesses. In contrast, TruncQuant reduces energy consumption by avoiding these expensive fetches in both the initial loading stage and subsequent precision adjustments. 
For the experiments in Fig.~\ref{fig:comp}, we leverage the Timeloop simulator’s built-in Eyeriss-like accelerator which employs row-stationary dataflow and hierarchical on-chip scratchpad and global buffers to minimize data movement~\cite{chen2016eyeriss}. We evaluate several scenarios in which TruncQuant dynamically switches model precision after the initial weight load, highlighting the reductions in energy consumption. In the legend, “MAC+spad” denotes the combined energy cost of the multiply–accumulate units and the input/output/weight scratchpad memories, while “glb\_buffer” refers to the energy consumption of the global buffer.

We also analyze TruncQuant’s hardware benefits in terms of storage savings. Enabling bit-switching through dedicated training increases storage requirements due to the need for multiple models. While once-for-all methods like Any-Precision eliminate this need, they still depend on the floating-point model to switch to lower precisions. TruncQuant, however, only stores the maximum bit precision within the switching range, demonstrating superior storage efficiency. This comparison is highlighted in Fig.~\ref{fig:storage}.

\section{Conclusion}
In this paper, we have proposed TruncQuant, which addresses the subtle but consequential differences between uniform quantization and truncation. 
We demonstrate that weights can be subject to changing bit precisions in runtime by using simple truncation, while preserving accuracy even in low bit precisions.
TruncQuant enables the deployment of neural networks that are capable of flexible operation in various real-world applications to adaptively achieve a balance between hardware efficiency and machine learning task accuracy. 

\section*{Acknowledgment}
This work was partly supported by the National Research Foundation of Korea (NRF) grant (No. RS-2024-00345732);
the Institute for Information \& communications Technology Planning \& Evaluation (IITP) grants (RS-2020-II201821, IITP-2021-0-02052, RS-2019-II190421, RS-2021-II212068); the Technology Innovation Program (RS-2023-00235718, 23040-15FC) funded by the Ministry of Trade, Industry \& Energy (MOTIE, Korea) grant (1415187505); Samsung Electronics Co., Ltd (IO230404-05747-01); and the BK21-FOUR Project.

\newpage
\bibliography{reference}

\begin{thebibliography}{10}
\providecommand{\url}[1]{#1}
\csname url@samestyle\endcsname
\providecommand{\newblock}{\relax}
\providecommand{\bibinfo}[2]{#2}
\providecommand{\BIBentrySTDinterwordspacing}{\spaceskip=0pt\relax}
\providecommand{\BIBentryALTinterwordstretchfactor}{4}
\providecommand{\BIBentryALTinterwordspacing}{\spaceskip=\fontdimen2\font plus
\BIBentryALTinterwordstretchfactor\fontdimen3\font minus \fontdimen4\font\relax}
\providecommand{\BIBforeignlanguage}[2]{{%
\expandafter\ifx\csname l@#1\endcsname\relax
\typeout{** WARNING: IEEEtran.bst: No hyphenation pattern has been}%
\typeout{** loaded for the language `#1'. Using the pattern for}%
\typeout{** the default language instead.}%
\else
\language=\csname l@#1\endcsname
\fi
#2}}
\providecommand{\BIBdecl}{\relax}
\BIBdecl

\bibitem{burrello2021dory}
A.~Burrello, A.~Garofalo, N.~Bruschi, G.~Tagliavini, D.~Rossi, and F.~Conti, ``Dory: Automatic end-to-end deployment of real-world dnns on low-cost iot mcus,'' \emph{IEEE Transactions on Computers}, vol.~70, no.~8, pp. 1253--1268, 2021.

\bibitem{yang2019quantization}
J.~Yang, X.~Shen, J.~Xing, X.~Tian, H.~Li, B.~Deng, J.~Huang, and X.-s. Hua, ``Quantization networks,'' in \emph{Proceedings of the IEEE/CVF Conference on Computer Vision and Pattern Recognition}, 2019, pp. 7308--7316.

\bibitem{banner2019post}
R.~Banner, Y.~Nahshan, and D.~Soudry, ``Post training 4-bit quantization of convolutional networks for rapid-deployment,'' \emph{Advances in Neural Information Processing Systems}, vol.~32, 2019.

\bibitem{nagel2020up}
M.~Nagel, R.~A. Amjad, M.~Van~Baalen, C.~Louizos, and T.~Blankevoort, ``Up or down? adaptive rounding for post-training quantization,'' in \emph{International Conference on Machine Learning}.\hskip 1em plus 0.5em minus 0.4em\relax PMLR, 2020, pp. 7197--7206.

\bibitem{li2021brecq}
Y.~Li, R.~Gong, X.~Tan, Y.~Yang, P.~Hu, Q.~Zhang, F.~Yu, W.~Wang, and S.~Gu, ``Brecq: Pushing the limit of post-training quantization by block reconstruction,'' \emph{arXiv preprint arXiv:2102.05426}, 2021.

\bibitem{yu2018slimmable}
J.~Yu, L.~Yang, N.~Xu, J.~Yang, and T.~Huang, ``Slimmable neural networks,'' \emph{arXiv preprint arXiv:1812.08928}, 2018.

\bibitem{yu2019universally}
J.~Yu and T.~S. Huang, ``Universally slimmable networks and improved training techniques,'' in \emph{Proceedings of the IEEE/CVF international conference on computer vision}, 2019, pp. 1803--1811.

\bibitem{tan2019efficientnet}
M.~Tan and Q.~Le, ``Efficientnet: Rethinking model scaling for convolutional neural networks,'' in \emph{International conference on machine learning}.\hskip 1em plus 0.5em minus 0.4em\relax PMLR, 2019, pp. 6105--6114.

\bibitem{teera2016branchynet}
S.~Teerapittayanon, B.~McDanel, and H.~T. Kung, ``{BranchyNet: Fast inference via early exiting from deep neural networks},'' in \emph{{Proceedings of the 23rd International Conference on Pattern Recognition}}.\hskip 1em plus 0.5em minus 0.4em\relax {IEEE}, 2016, pp. 2464--2469.

\bibitem{laskaridis2021adaptive}
S.~Laskaridis, A.~Kouris, and N.~D. Lane, ``Adaptive inference through early-exit networks: Design, challenges and directions,'' in \emph{Proceedings of the 5th International Workshop on Embedded and Mobile Deep Learning}, 2021, pp. 1--6.

\bibitem{ryu2022review}
S.~Ryu, ``Review and analysis of variable bit-precision mac microarchitectures for energy-efficient ai computation,'' \emph{JOURNAL OF SEMICONDUCTOR TECHNOLOGY AND SCIENCE}, vol.~22, no.~5, pp. 353--360, 2022.

\bibitem{camus2019review}
V.~Camus, L.~Mei, C.~Enz, and M.~Verhelst, ``Review and benchmarking of precision-scalable multiply-accumulate unit architectures for embedded neural-network processing,'' \emph{IEEE Journal on Emerging and Selected Topics in Circuits and Systems}, vol.~9, no.~4, pp. 697--711, 2019.

\bibitem{sharma2018bit}
H.~Sharma, J.~Park, N.~Suda, L.~Lai, B.~Chau, J.~K. Kim, V.~Chandra, and H.~Esmaeilzadeh, ``Bit fusion: Bit-level dynamically composable architecture for accelerating deep neural network,'' in \emph{2018 ACM/IEEE 45th Annual International Symposium on Computer Architecture (ISCA)}.\hskip 1em plus 0.5em minus 0.4em\relax IEEE, 2018, pp. 764--775.

\bibitem{moons2017envision}
B.~Moons, R.~Uytterhoeven, W.~Dehaene, and M.~Verhelst, ``14.5 envision: A 0.26-to-10tops/w subword-parallel dynamic-voltage-accuracy-frequency-scalable convolutional neural network processor in 28nm fdsoi,'' in \emph{2017 IEEE International Solid-State Circuits Conference (ISSCC)}, 2017, pp. 246--247.

\bibitem{ryu2019bitblade}
S.~Ryu, H.~Kim, W.~Yi, and J.-J. Kim, ``Bitblade: Area and energy-efficient precision-scalable neural network accelerator with bitwise summation,'' in \emph{Proceedings of the 56th Annual Design Automation Conference 2019}, 2019, pp. 1--6.

\bibitem{tahmasebi2024flexibit}
F.~Tahmasebi, Y.~Wang, B.~Y. Huang, and H.~Kwon, ``Flexibit: Fully flexible precision bit-parallel accelerator architecture for arbitrary mixed precision ai,'' \emph{arXiv preprint arXiv:2411.18065}, 2024.

\bibitem{xu2023eq}
K.~Xu, L.~Han, Y.~Tian, S.~Yang, and X.~Zhang, ``Eq-net: Elastic quantization neural networks,'' in \emph{Proceedings of the IEEE/CVF International Conference on Computer Vision}, 2023, pp. 1505--1514.

\bibitem{chmiel2020robust}
B.~Chmiel, R.~Banner, G.~Shomron, Y.~Nahshan, A.~Bronstein, U.~Weiser \emph{et~al.}, ``Robust quantization: One model to rule them all,'' \emph{Advances in neural information processing systems}, vol.~33, pp. 5308--5317, 2020.

\bibitem{xu2022multiquant}
K.~Xu, Q.~Feng, X.~Zhang, and D.~Wang, ``Multiquant: Training once for multi-bit quantization of neural networks.'' in \emph{IJCAI}, 2022, pp. 3629--3635.

\bibitem{zhong2023mbquant}
Y.~Zhong, Y.~Zhou, F.~Chao, and R.~Ji, ``Mbquant: A novel multi-branch topology method for arbitrary bit-width network quantization,'' \emph{arXiv preprint arXiv:2305.08117}, 2023.

\bibitem{dao2022flashattention}
T.~Dao, D.~Y. Fu, S.~Ermon, A.~Rudra, and C.~R{\'e}, ``Flash{A}ttention: Fast and memory-efficient exact attention with {IO}-awareness,'' in \emph{Advances in Neural Information Processing Systems}, 2022.

\bibitem{yu2021any}
H.~Yu, H.~Li, Haoxiang opand~Shi, T.~S. Huang, and G.~Hua, ``Any-precision deep neural networks,'' in \emph{Proceedings of the AAAI Conference on Artificial Intelligence}, vol.~35, no.~12, 2021, pp. 10\,763--10\,771.

\bibitem{gholami2022survey}
A.~Gholami, S.~Kim, Z.~Dong, Z.~Yao, M.~W. Mahoney, and K.~Keutzer, ``A survey of quantization methods for efficient neural network inference,'' in \emph{Low-Power Computer Vision}.\hskip 1em plus 0.5em minus 0.4em\relax Chapman and Hall/CRC, 2022, pp. 291--326.

\bibitem{qin2020binary}
H.~Qin, R.~Gong, X.~Liu, X.~Bai, J.~Song, and N.~Sebe, ``Binary neural networks: A survey,'' \emph{Pattern Recognition}, vol. 105, p. 107281, 2020.

\bibitem{jacob2018quantization}
B.~Jacob, S.~Kligys, B.~Chen, M.~Zhu, M.~Tang, A.~Howard, H.~Adam, and D.~Kalenichenko, ``Quantization and training of neural networks for efficient integer-arithmetic-only inference,'' in \emph{Proceedings of the IEEE conference on computer vision and pattern recognition}, 2018, pp. 2704--2713.

\bibitem{esser2019learned}
S.~K. Esser, J.~L. McKinstry, D.~Bablani, R.~Appuswamy, and D.~S. Modha, ``Learned step size quantization,'' \emph{arXiv preprint arXiv:1902.08153}, 2019.

\bibitem{shen2021once}
M.~Shen, F.~Liang, R.~Gong, Y.~Li, C.~Li, C.~Lin, F.~Yu, J.~Yan, and W.~Ouyang, ``Once quantization-aware training: High performance extremely low-bit architecture search,'' in \emph{Proceedings of the IEEE/CVF International Conference on Computer Vision}, 2021, pp. 5340--5349.

\bibitem{ste}
Y.~Bengio, N.~L{\'e}onard, and A.~Courville, ``Estimating or propagating gradients through stochastic neurons for conditional computation,'' \emph{arXiv preprint arXiv:1308.3432}, 2013.

\bibitem{zhou2016dorefa}
S.~Zhou, Y.~Wu, Z.~Ni, X.~Zhou, H.~Wen, and Y.~Zou, ``Dorefa-net: Training low bitwidth convolutional neural networks with low bitwidth gradients,'' \emph{arXiv preprint arXiv:1606.06160}, 2016.

\bibitem{liu2022nonuniform}
Z.~Liu, K.-T. Cheng, D.~Huang, E.~P. Xing, and Z.~Shen, ``Nonuniform-to-uniform quantization: Towards accurate quantization via generalized straight-through estimation,'' in \emph{Proceedings of the IEEE/CVF Conference on Computer Vision and Pattern Recognition}, 2022, pp. 4942--4952.

\bibitem{yin2019understanding}
P.~Yin, J.~Lyu, S.~Zhang, S.~Osher, Y.~Qi, and J.~Xin, ``Understanding straight-through estimator in training activation quantized neural nets,'' \emph{arXiv preprint arXiv:1903.05662}, 2019.

\bibitem{he2016deep}
K.~He, X.~Zhang, S.~Ren, and J.~Sun, ``Deep residual learning for image recognition,'' in \emph{Proceedings of the IEEE conference on computer vision and pattern recognition}, 2016, pp. 770--778.

\bibitem{sandler2018mobilenetv2}
M.~Sandler, A.~Howard, M.~Zhu, A.~Zhmoginov, and L.-C. Chen, ``Mobilenetv2: Inverted residuals and linear bottlenecks,'' in \emph{Proceedings of the IEEE conference on computer vision and pattern recognition}, 2018, pp. 4510--4520.

\bibitem{krizhevsky2012imagenet}
A.~Krizhevsky, I.~Sutskever, and G.~E. Hinton, ``Imagenet classification with deep convolutional neural networks,'' \emph{Advances in neural information processing systems}, vol.~25, 2012.

\bibitem{krizhevsky2009learning}
A.~Krizhevsky, G.~Hinton \emph{et~al.}, ``Learning multiple layers of features from tiny images,'' 2009.

\bibitem{deng2009imagenet}
J.~Deng, W.~Dong, R.~Socher, L.-J. Li, K.~Li, and L.~Fei-Fei, ``Imagenet: A large-scale hierarchical image database,'' in \emph{2009 IEEE conference on computer vision and pattern recognition}.\hskip 1em plus 0.5em minus 0.4em\relax Ieee, 2009, pp. 248--255.

\bibitem{netzer2011reading}
Y.~Netzer, T.~Wang, A.~Coates, A.~Bissacco, B.~Wu, A.~Y. Ng \emph{et~al.}, ``Reading digits in natural images with unsupervised feature learning,'' in \emph{NIPS workshop on deep learning and unsupervised feature learning}, vol. 2011, no.~5.\hskip 1em plus 0.5em minus 0.4em\relax Granada, Spain, 2011, p.~7.

\bibitem{parashar2019timeloop}
A.~Parashar, P.~Raina, Y.~S. Shao, Y.-H. Chen, V.~A. Ying, A.~Mukkara, R.~Venkatesan, B.~Khailany, S.~W. Keckler, and J.~Emer, ``Timeloop: A systematic approach to dnn accelerator evaluation,'' in \emph{2019 IEEE international symposium on performance analysis of systems and software (ISPASS)}.\hskip 1em plus 0.5em minus 0.4em\relax IEEE, 2019, pp. 304--315.

\bibitem{chen2016eyeriss}
Y.-H. Chen, T.~Krishna, J.~S. Emer, and V.~Sze, ``Eyeriss: An energy-efficient reconfigurable accelerator for deep convolutional neural networks,'' \emph{IEEE journal of solid-state circuits}, vol.~52, no.~1, pp. 127--138, 2016.

\end{thebibliography}

\end{document}